\let\texyear\year

\documentclass{ieeeaccess}
\usepackage{cite}
\usepackage{amsmath,amssymb,amsfonts}
\usepackage{pifont}
\usepackage{algorithm,algorithmic}
\usepackage{multirow}
\usepackage{diagbox}
\usepackage{array}
\usepackage{url}
\usepackage[pagebackref=true,breaklinks=true,colorlinks,bookmarks=false]{hyperref}
\usepackage{graphicx}
\usepackage{textcomp}

\let\ieeeaccessyear\year
\let\year\texyear
    \usepackage{orcidlink}
\let\year\ieeeaccessyear
\definecolor{accessblue}{RGB}{0,105,154}

\newcommand{\etal}{\textit{et al. }}
\newcommand{\xmark}{\ding{55}}

\def\BibTeX{{\rm B\kern-.05em{\sc i\kern-.025em b}\kern-.08em
    T\kern-.1667em\lower.7ex\hbox{E}\kern-.125emX}}
\begin{document}
\history{Date of publication xxxx 00, 0000, date of current version xxxx 00, 0000.}
\doi{10.1109/ACCESS.2022.DOI}

\title{LAD: A Hybrid Deep Learning System for Benign Paroxysmal Positional Vertigo Disorders Diagnostic}

% \author{\uppercase{First A. Author}\authorrefmark{1}, \IEEEmembership{Fellow, IEEE},
% \uppercase{Second B. Author\authorrefmark{2}, and Third C. Author,
% Jr}.\authorrefmark{3},
% \IEEEmembership{Member, IEEE}}
% \address[1]{National Institute of Standards and 
% Technology, Boulder, CO 80305 USA (e-mail: author@boulder.nist.gov)}
% \address[2]{Department of Physics, Colorado State University, Fort Collins, 
% CO 80523 USA (e-mail: author@lamar.colostate.edu)}
% \address[3]{Electrical Engineering Department, University of Colorado, Boulder, CO 
% 80309 USA}
% \tfootnote{This paragraph of the first footnote will contain support 
% information, including sponsor and financial support acknowledgment. For 
% example, ``This work was supported in part by the U.S. Department of 
% Commerce under Grant BS123456.''}

\author{\uppercase{Trung Xuan Pham}$^{\orcidlink{0000-0003-4177-7054}}$\authorrefmark{1}$^{,\dagger}$, \IEEEmembership{Student Member, IEEE},
% corrected here
\uppercase{Jin Woong Choi$^{\orcidlink{0000-0003-3101-6841}}$\authorrefmark{2}${^{,\dagger}}$ \IEEEmembership{Senior Member, IEEE}}
\uppercase{Rusty John Lloyd Mina}\authorrefmark{1}, \IEEEmembership{Student Member, IEEE},
\uppercase{Thanh Nguyen}\authorrefmark{1}, \IEEEmembership{Student Member, IEEE},
\uppercase{Sultan Rizky Madjid}\authorrefmark{1}, \IEEEmembership{Student Member, IEEE},
\uppercase{Chang Dong Yoo}$^{\orcidlink{0000-0002-0756-7179}}$\authorrefmark{1}$^{,*}$, \IEEEmembership{Senior Member, IEEE}}

\address[1]{School of Electrical Engineering, Korea Advanced Institute of Science and Technology (KAIST). Address: KAIST, 291 Daehak-ro, 373-1 Guseong Dong, Yuseong Gu, Daejeon, 305-701, zip code: 34141, Republic of Korea (e-mail: trungpx@kaist.ac.kr)}

% \address[2]{Chungnam National University (CNU), Department of Otolaryngology-Head and Neck Surgery. Address: 99 Daehak-ro, Yuseong-gu, Daejeon, Republic of Korea (e-mail: choijw@cnu.ac.kr)}
% corrected here
\address[2]{Department of Otorhinolaryngology-Head and Neck Surgery, Chungnam National University, School of Medicine. Address: 282 Munhwa-ro, Jung-gu, Daejeon, 301-721, zip code: 35015, Republic of Korea (e-mail: choijw@cnu.ac.kr)}

% \tfootnote{This work was supported in part by the Institute for Information \& communications Technology Promotion (IITP) grant funded by the Korean government (MSIT) (No. 2021-0-01381, Development of Causal AI through Video Understanding) and in part by the Institute of Information \& communications Technology Planning \& Evaluation (IITP) grant funded by the Korean government (MSIT) (2022-0-00951, Development of Uncertainty-Aware Agents Learning by Asking Questions.}
\tfootnote{This work was supported in part by the Institute for Information \& communications Technology Promotion (IITP) grant funded by the Korean government (MSIT) (No. 2021-0-01381, Development of Causal AI through Video Understanding) and was partly supported by the National Research Foundation of Korea (NRF) grant funded by the Korea government (MSIT) (No. 2022R1A2C201270611).} %corrected here

\markboth
{T.X.Pham \headeretal: Preparation of Papers for IEEE TRANSACTIONS and JOURNALS}
{T.X.Pham \headeretal: Preparation of Papers for IEEE TRANSACTIONS and JOURNALS}

% corrected here
\corresp{$^*$Corresponding author: Chang D. Yoo (e-mail: cd\_yoo@kaist.ac.kr). $^{\dagger}$The first two authors contributed equally to this study.}

\begin{abstract}
    Herein, we introduce ``Look and Diagnose'' (LAD), a hybrid deep learning-based system that aims to support doctors in the medical field for diagnosing effectively the \textit{Benign Paroxysmal Positional Vertigo} (BPPV) disorder. Given the body postures of the patient in the Dix-Hallpike and lateral head turns test, the visual information of both eyes is captured and fed into LAD for analyzing and classifying into one of six possible disorders which the patient might be suffering from. The proposed system consists of two streams: (1) an RNN-based stream that takes raw RGB images of both eyes to extract visual features and optical flow of each eye followed by ternary classification to determine left/right posterior canal (PC) or other; and (2) pupil detector stream that detects the pupil when it is classified as Non-PC and classifies the direction and strength of the beating to categorize the Non-PC types into the remaining four classes: \textit{Geotropic} BPPV (left and right) and \textit{Apogeotropic} BPPV (left and right). Experimental results show that with the given body postures of the patient, the system is capable of accurately classifying given BPPV disorder into the six types of disorder with an accuracy of 91\% on the validation set. The proposed method can successfully classify disorders with an accuracy of 93\% for the \textit{Posterior Canal} disorder and 95\% for the \textit{Geotropic} and \textit{Apogeotropic} disorder, paving a potential direction for research with the medical data.
\end{abstract}


\titlepgskip=-15pt

\maketitle

\section{Introduction}
\label{sec:introduction}

Benign Paroxysmal Positional Vertigo is one of the most common causes of vertigo with a disorder arising from the problem in the inner ear \cite{pengyou}. Symptoms are repeated, brief periods of vertigo with movement, characterized by a spinning sensation upon changes in the position of the head \cite{bhatta}. Each episode lasts less than one minute and nausea is commonly associated.
    Each part of the name depicts a key feature of the inner-ear disorder: \textit{Benign} means it is not very serious, the patient's life is not in danger; \textit{Paroxysmal} indicates that the symptom occurs suddenly and takes place within a short time; \textit{Positional} implies that the vertigo symptom is triggered for certain postures and head movements.
    %The vertigo is triggered for certain postures and head movements of the patient.
    BPPV is considered a common public health problem \cite{yetiser}. The probable causes of BPPV are different lesions such as head trauma, head injury, or some surgeries related to otology, oral, and maxillofacial cases \cite{riga}. The dizziness symptom of BPPV is explained by loose calcium deposits (crystals or “ear rocks”) in what is called the semicircular canals of the inner ear \cite{infohealth}. When the patient's head is moved, these crystals roll around the semicircular canal, which transmits misleading information (like what the eyes are seeing) with the information received by the brain and these conflicting signals make the person's dizziness happens. There are around 5 to 6 million people who suffer from this disorder annually \cite{liudh,balat,byun}. In some cases, the problem of BPPV might become serious when it causes increasing chances of falling and losing balance for patients \cite{horni}. The abnormal movements of the eyes help the doctor determine the type of BPPV disorder. The symptoms of benign paroxysmal positional vertigo usually often accompany abnormal rhythmic eye movements. These behaviors include nystagmus (beating) and torsional motions of the eyes. Based on the different nystagmus movements of the eye, the doctor is able to categorize BPPV into different disorders for the appropriate treatments. Differentiating different disorders is a challenging task as it requires expert knowledge and careful observation of the patient. BPPV disorder can occur in all age groups and gender. However, it is rare for someone under 20 years to have the disorder but quite common for someone in the middle age group of 31-50 years \cite{yetiser}. 
   \begin{figure}[!ht]
    	\centering
    	\includegraphics[width=1.0\linewidth]{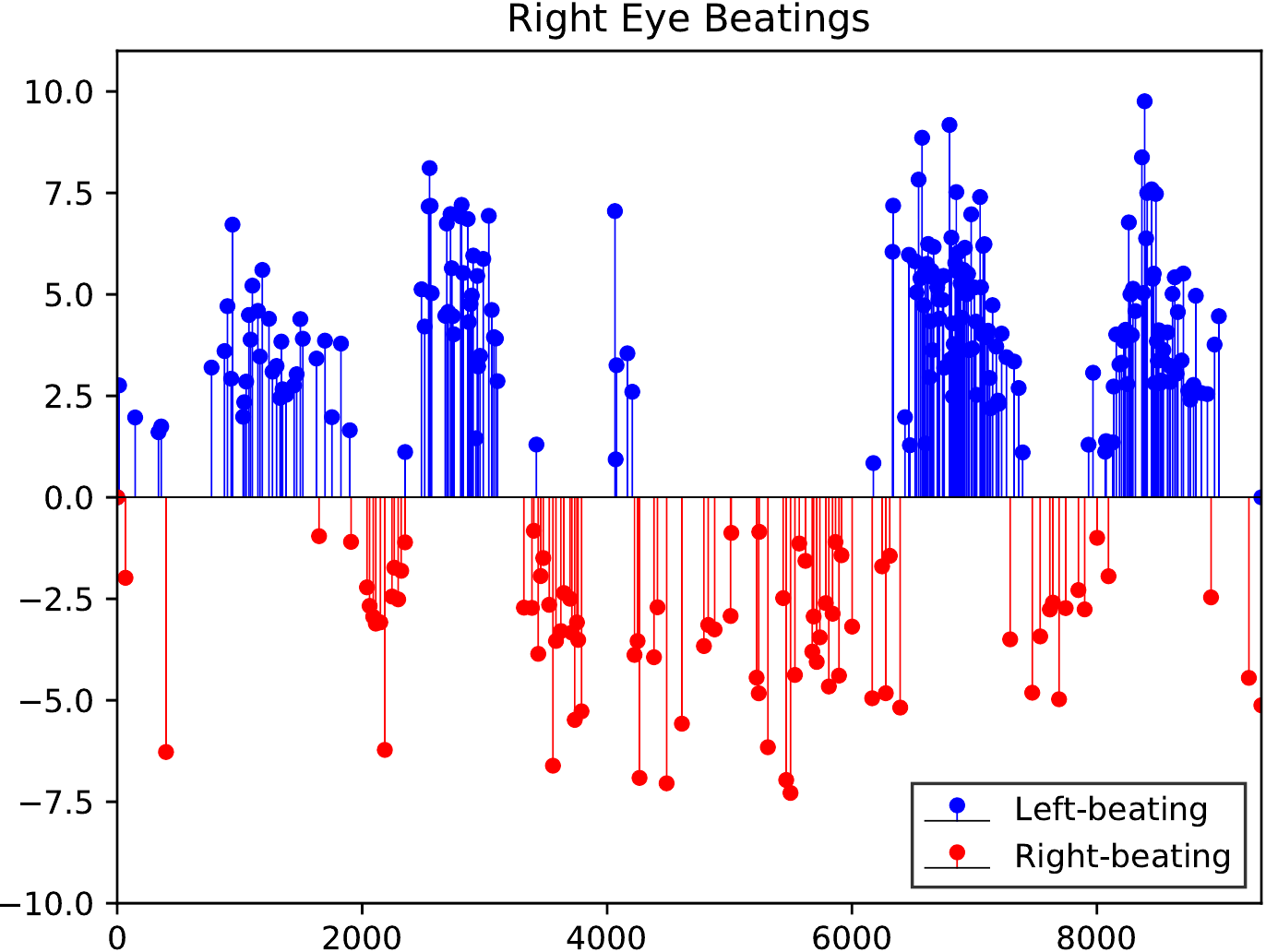}
    	\caption{Horizontal beating detection of the right eye has been analyzed for a patient who is diagnosed to be positive to type Lt\_Geo\_BPPV with 9676 video frames. Red points are the right beating, blue points are left beatings. The horizontal axis is the number of frames, the vertical axis is velocity.}
    	\label{fig:fig01_right}
    \end{figure}
    
    BPPV is often benign, however, it may be dangerous in certain cases. For example, when a BPPV patient is working on a ladder or on the top roof, he may suddenly be hit by a vertiginous symptom and he may lose balance which may lead to an unwanted serious accident. If the kind of disorder is accurately diagnosed, prompt treatment can be applied so that the patient can be cured earlier and avoid the painful experience and future risks \cite{tangheng}. BPPV disorders are diagnosed by observing the visual motions of the human eyes in some clinical settings \cite{kerber,dimitris,lee2019clinical}. % corrected here, add 1 reference
       \begin{figure}[!ht]
    	\centering
    	\includegraphics[width=1.0\linewidth]{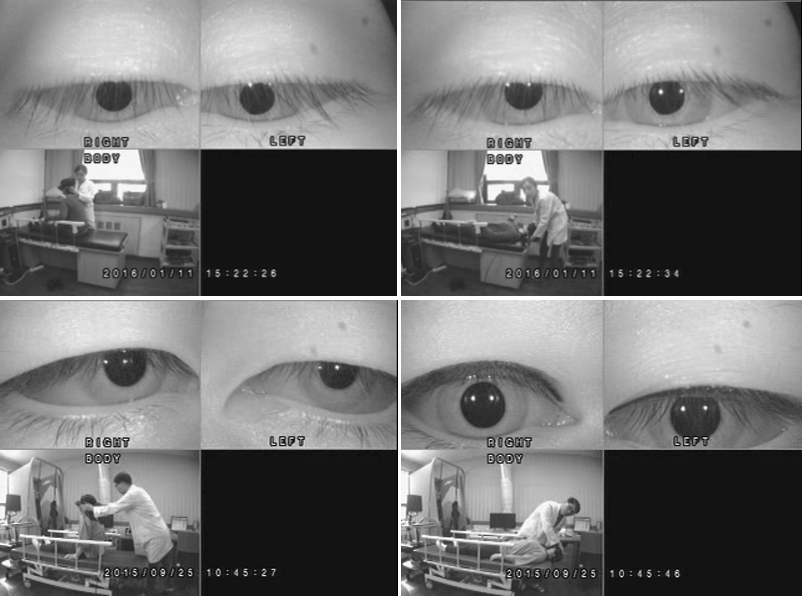}
    	\caption{Four examples of the collected video dataset. Following the Dix-Hallpike test, \cite{tangheng}, different poses of the patient body were conducted for the diagnosis of BPPV disorders with the visual observation corresponding to the eye's abnormal moving. Each video contains 3 sub-videos which record the left eye, right eye, and body pose. The black area does not contain information.}
    	\label{fig:dataset}
    \end{figure}
    \begin{figure*}[!ht]
    	\centering
    	\includegraphics[width=1\linewidth]{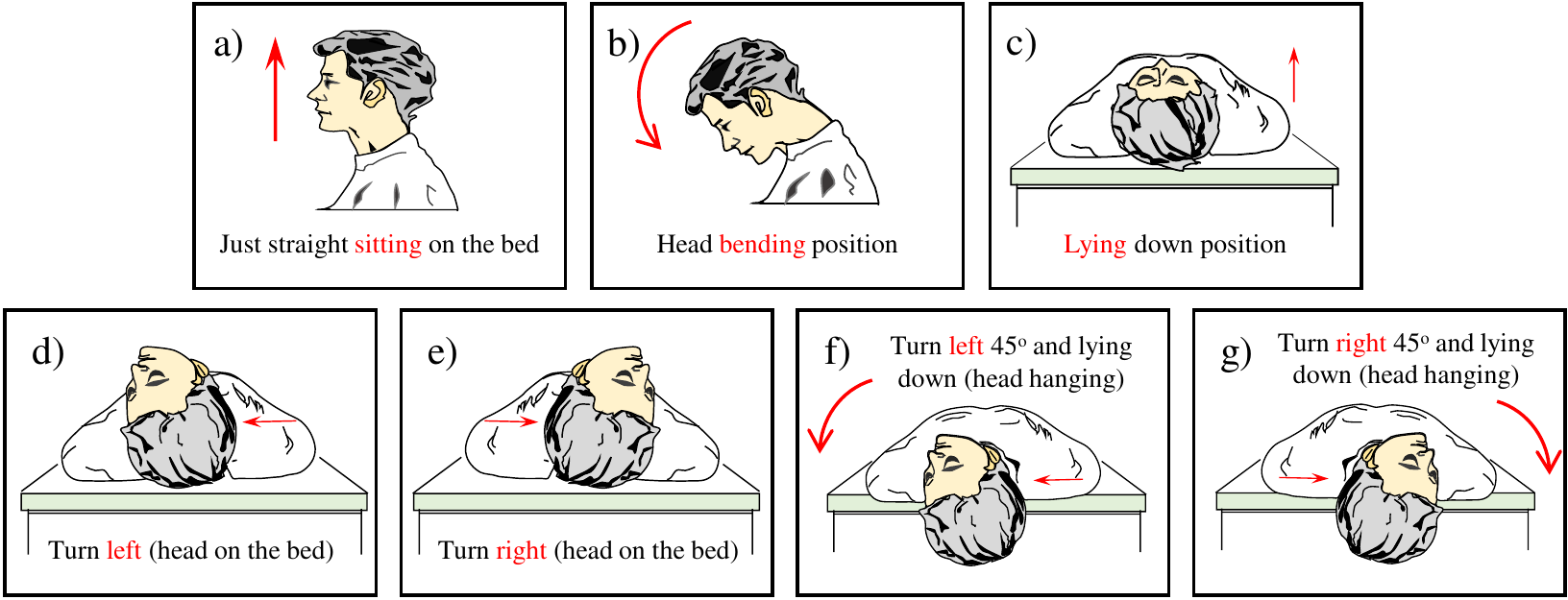}
    	\caption{A normal procedure of BPPV diagnosis which collects video data of the eyes associated with the seven positions of the patient's body: (a) the patient is sitting straight, (b) head bending while sitting position, (c) laying down while patient keeps looking straightly into the ceiling, (d) when patient laid on the bed, perform turning head left quickly, (e) when patient laid on the bed try turning head right quickly, (f) from the sitting position, try turning head left $ 45^\text{o} $ lying down and make sure the head is hanging out of the bed, and (g) from the sitting position, turning head right $ 45^\text{o} $ lying down and make sure the head is hanging out of the bed. Performing actions \textit{b, c, d, e} often indicates the ``lateral head turns'' test while the action \textit{a, f, g} are intended the ``Dix-Hallpike'' test.}
    	\label{fig:fig05}
    \end{figure*}
    
    Based on any abnormal movements of the eyes, the otology experts are able to categorize each case to a specific type for further appropriate treatments. When the number of patients increases, it is a burden for medical employees. The cost of BPPV diagnosis is estimated to be approximately \$2000, and most patients suffering from this disorder may undergo unnecessary testing or other interventions which leads to the associated cost of billions of dollars per year for diagnosing BPPV alone \cite{bhatta}. Deep learning has made a huge impact in computer vision that includes image classification \cite{resnet,tan19a,xie2020}, video question and answering \cite{tvqa,Kim_2020_CVPR, donahue}, object recognition \cite{crpn}, and natural language processing \cite{bert}. It is helping to identify, classify, and quantify patterns in medical images. 
    In \cite{ryan}, Inception V3 is used to extract new knowledge from retinal fundus images and infer multiple cardiovascular risk factors with the high area under the receiver operating characteristic curve (AUC). In \cite{andre}, the fine-grained variability in the appearance of skin lesions has been effectively captured by using CNN with manipulated transfer learning to detect skin cancer and achieve performance on par with the clinical experts. Recent studies demonstrate the potential applications of deep learning for understanding human actions in the videos \cite{carrei,NIPS2014_00ec53c4,singh,rama}. These researches show the promising capability of deep learning in capturing the torsional nystagmus patterns in the fine-grained motion of the eyes for BPPV analysis. 
    \begin{figure}[!ht] %ht
    	\centering
    	\includegraphics[width=1.0\linewidth]{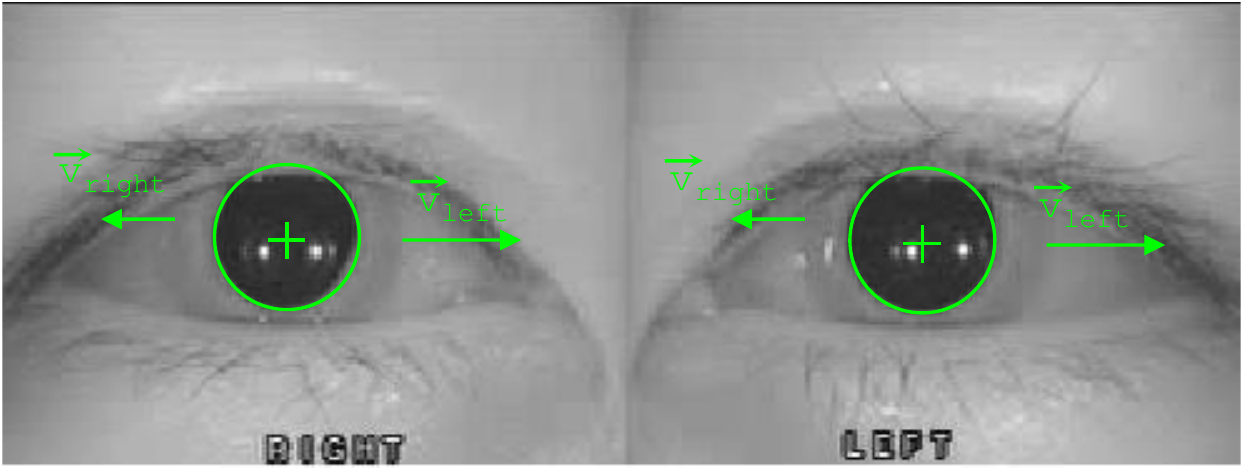}
    	\caption{A sample of our video data. BPPV disorder is specified by the natural nystagmus of the people's eyes along with the featured motions such as torsional and spontaneous beating movements. Horizontal beatings are measured and detected by the proposed algorithm to diagnose the lateral canal BPPV types, while torsional motions are captured by the RNN (GRU) network.}
    	\label{fig:fig03}
    \end{figure}
    In this paper, a deep learning system referred to as "Look and Diagnose" (LAD) is capable of classifying BPPV disorder into six types of the most common BPPVs including a geotropic lateral canal (left and right), apogeotropic lateral canal (left and right), and posterior canal types (left and right) based on the video of both eyes and seven postures of the patient. LAD is capable of classifying different types of beating (nystagmus) including torsional and horizontal beating to effectively classify BPPV disorders. To develop LAD, we have collected a large dataset of both eye movement videos of BPPV disorders. More details are shown in section \ref{sec:experiments}.

\section{Related Works}
\label{relatedworks}
    \subsection{BPPV analysis}
    Nystagmus examination is a crucial step to diagnose BPPV disorder \cite{pengyou}. Peng \etal investigated different maneuvers to provoke the nystagmus for diagnosis and the BPPV can be treated effectively by applying repositional moving \cite{tangheng}. Previous studies show that horizontal (lateral) canal type (5\% to 15\% of BPPV cases) is less common than the Posterior canal type (85\% to 95\% of BPPV cases) \cite{bhatta}. 
    
    Slama \etal \cite{slama2,slama3} used a multilayer neural network (MNN) and the recorded parameters from Video Nysta Geographic (VNG) data to analyze the nystagmus to diagnose whether one person has a vestibular disorder or normal. Nystagmus signal from Videonystagmography (VNG) device has been analyzed by CNN-based method to classify two classes of vestibular disorder \cite{slama4}. Lim \etal \cite{limec} introduced a more complete approach that does not just only classify the nystagmus, but also diagnoses the final BPPV class using various ad-hoc techniques. 
    
    Lim \etal recorded the tracking videos of eyes at ten different postures of the patient. These videos have been processed to get the cumulative transient velocity of the eye movement by simply subtracting eye coordinates for every video frame. The extracted scalars have been formed into a grid image for each video and fed into a standard CNN classifier. Lim \etal is based on hand-crafted features for diagnosing BPPV disorder, which has limited their performance. 
    
    In contrast to \cite{limec}, we collected videos from our clinical warehouse with seven postures of patient (Fig. \ref{fig:fig05}, Fig. \ref{fig:dataset}). A bidirectional GRU-based network was used to successfully learn end-to-end the torsional movements and nystagmus simultaneously of the two eyes from the given video input. % corrected here, add Fig fig:dataset
    
    A consecutive feature-based classifier captures the horizontal beating of the eyes and recognizes the remaining four horizontal (lateral) canal BPPV disorder classes including geotropic BPPV (left and right side) and apogeotropic BPPV (left and right). 
    
    \subsection{Action recognition}
    Video understanding is an attractive and challenging task. Some earlier works tried to extract the handcrafted features of the video via improved Dense Trajectory (iDT) and combine it with low-level video descriptors for video representation \cite{idt,idt2,carrei}. Deep learning has made a significant impact on human action recognition \cite{access9857832,access9807301,access9802089,access9802105,access9779234,access9652524,access9631267}. Two-stream networks are capable of extracting relevant features regarding the appearance and motion information from RGB and optical flow field for accurate action recognition \cite{NIPS2014_00ec53c4,rama,tvnet}. 
    
    Another approach for action recognition uses the 3D-CNN networks to directly capture motion clues in the video by using 3D convolution without using optical flow field \cite{c3d,dutran}. In this work, a two-stream system is adopted to extract the motion features such as the torsional nystagmus of the eyes for accurate disorders classification.
    
    \subsection{Eye pupil tracking}
    Traditional computer vision methods use edge features, intensity gradient distribution and intensity thresholds to detect the eye pupil \cite{vtm,hog}. A more advanced feature-based technique uses edge segment selection and conditional segment combination for pupil detection with high accuracy \cite{pure}. Recently, a deep learning-based network is incorporated to robustly detect pupils even under extreme conditions with reflections and occlusion by eyelids \cite{deepeye,shahara}. 
    
    In this work, the pre-trained CNN model for localizing the pupil proposed by \cite{shahara} is adapted, and the pupil position is monitored in the proposed system for classifying horizontal beating so that four types of lateral canal BPPV disorders are categorized. To this end, an algorithm is proposed to process the pupil location to detect when the beating happened and measure the speed of horizontal beating for lateral canal BPPV.
    
    \begin{figure}[!ht]
    	\centering
    	\includegraphics[width=1.0\linewidth]{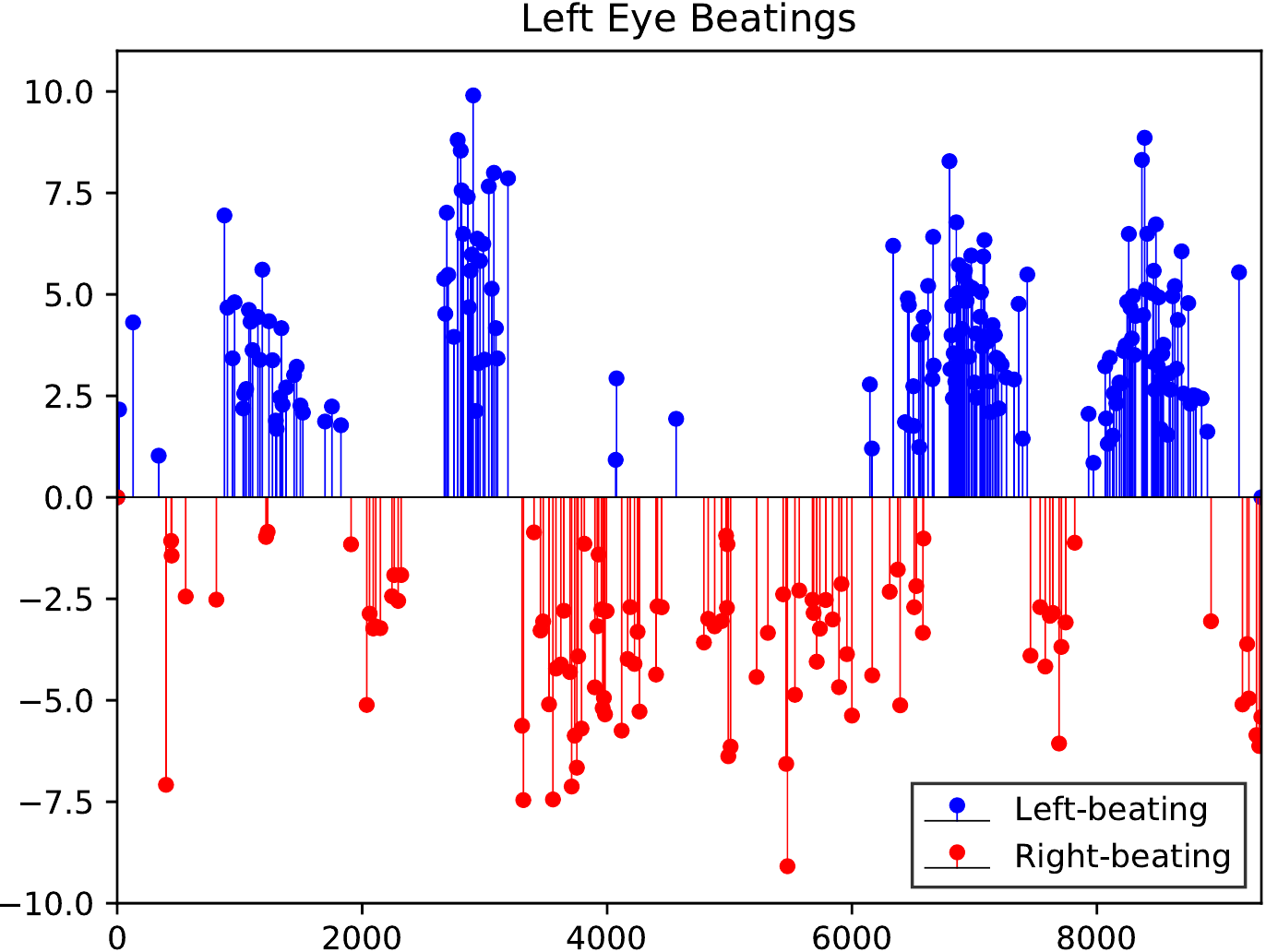}
    	\caption{Horizontal beating detection of the left eye has been analyzed for a patient who is diagnosed to be positive to type Lt\_Geo\_BPPV with 9676 video frames. Red points are the right beating, blue points are left beatings. The horizontal axis is the number of frames, the vertical axis is velocity.}
    	\label{fig:fig01}
    \end{figure}
    
\section{Methodology}
    \label{sec:methodology}
    This section describes the details of the proposed  BPPV disorder diagnosis system referred to as LAD which takes as input the video recording of eye movement of both eyes with posture labels. Seven postures labels aligned with the eye movement are provided. Fig. \ref{fig:fig05} shows the seven postures that include sitting, head on the bed turned left, head on the bed turned right, head hanging $ 45^\text{o} $ to the left, head hanging $ 45^\text{o} $ to the right, lying down position, and head bending a sitting position. 
    
    Based on these positions, the system learns to capture the relevant features of both eyes from RGB video and optical flow output of the optical flow CNN and based on a bi-directional gated recurrent unit (BiGRU) framework categorizes the input video into two types of posterior canal BPPV (PC) and other (not PC).
    \begin{figure}[!ht]
    	\centering
    	\includegraphics[width=1.0\linewidth]{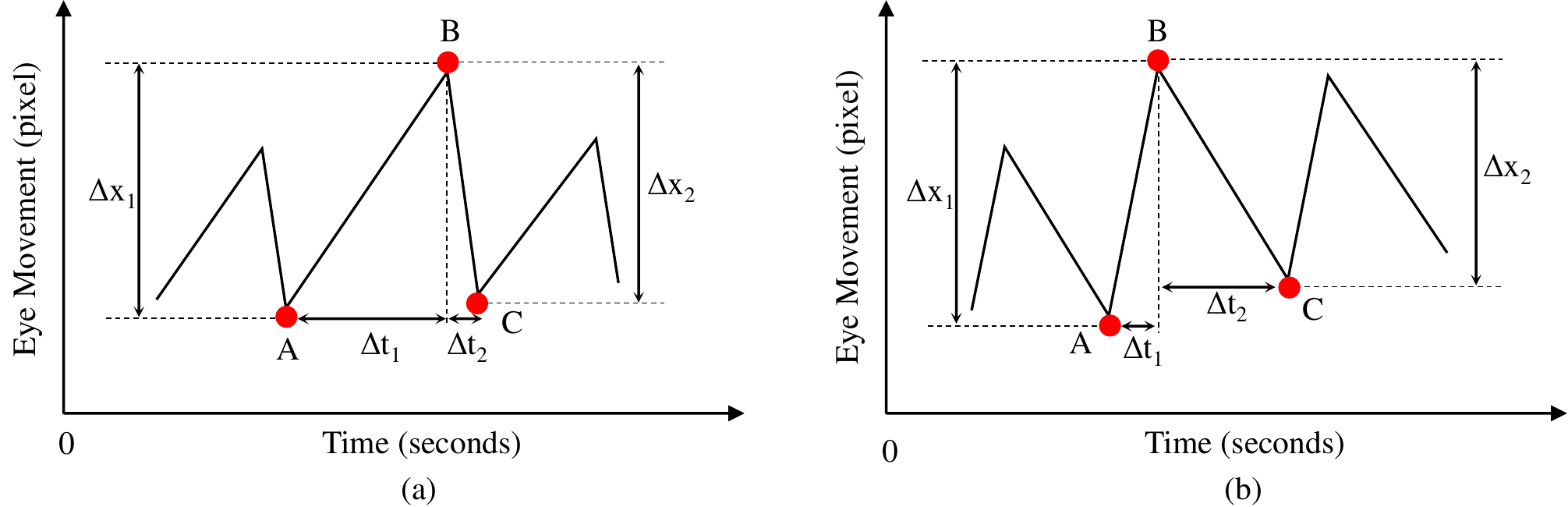}
    	\caption{The horizontal beating is determined by slow phase velocity (SPV) and fast phase velocity (FPV). Fig. (a) illustrates the eye has left-beating. In this case, the eye move from point B to point C faster (decreasing the pixel) than the eye moves from A to B. Whilst Fig. (b) shows the right-beating behavior of the eyes where they move from point A to point B faster than when it moves from point B to point C. The behavior of these beatings is contradicted. Here, $ \Delta x_1 $ indicates the distance when the eye travels during the AB in the direction of pixel increasing and $ \Delta x_2 $ is the distance when the eyes move toward the decreasing pixel direction. Similarly, we can determine the beating up and beating down where the eyes have the beating direction vertical. The beating is counted if the beating velocity is above a threshold.}
    	\label{fig:fig06}
    \end{figure}
    If the video is classified as not PC type, then the beating feature of the eye is extracted to categorize it into one of four remaining BPPV disorder types.
    \begin{algorithm}
    	\caption{Beating Detection (\textbf{pseudo code}), Fig. \ref{fig:fig06}}
    	\begin{algorithmic}[1]
    		\STATE \textbf{Input:} Detected \textbf{x} coordinates of pupil.
    		\STATE \textbf{Output:} Beating velocity, frame ID beating occurred.
    		\STATE Initialization: $counter_1 \gets 1$, $counter_2 \gets 1$, $beat_{left} \gets \text{NULL}$, $beat_{right} \gets \text{NULL}$
    		\FOR {frame i=2:N}
    		    \STATE $flag_1 \gets$ sign(\textbf{x}[i-1]-\textbf{x}[i-2])
    		    \STATE $flag_2 \gets$ sign(\textbf{x}[i]-\textbf{x}[i-1])
    		    \IF{$flag_1>0$ and $flag_2>0$}
        		\STATE Increase $counter_1$
        		\ELSIF{$flag_1<0$ and $flag_2<0$}
        		\STATE Increase $counter_2$
        		\ELSIF{$flag_1>0$ and $flag_2<0$}
        		\STATE Calculate slow phase velocity: $v_{left} \gets \frac{\Delta x}{counter_1}$
        		\ELSE
        		\STATE Calculate fast phase velocity: $v_{right} \gets \frac{\Delta x}{counter_2}$
        		\ENDIF
        		\STATE \textbf{if} $v_{left}<v_{right}$ \textbf{then} store $beat_{right}$ \textbf{else} store $beat_{left}$, reset counters to 1.
    		\ENDFOR
    	\end{algorithmic}
    \end{algorithm}
    
    \begin{figure*}[!ht]
    	\centering
    	\includegraphics[width=1.0\linewidth]{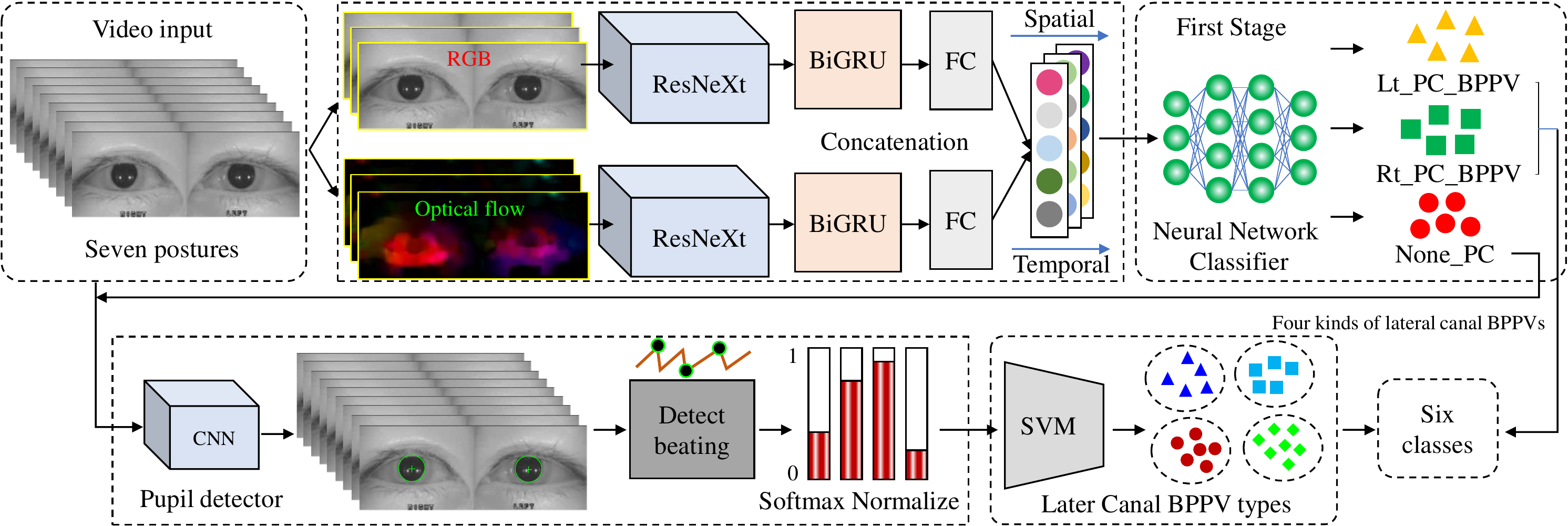}
    	\caption{The whole architecture for detecting six BPPV disorders. In the first stage, the input video is categorized into three raw classes ``posterior types'' and ``not posterior types''. After that, a pupil detector based on CNN is applied to track the eye pupils and detect the beating that happens in the second stage.}
    	\label{fig:fig04}
    \end{figure*}
    
    \subsection{Classifying Posterior and Lateral Types}
\begin{figure}[!ht]
    	\centering
    	\includegraphics[width=0.9\linewidth]{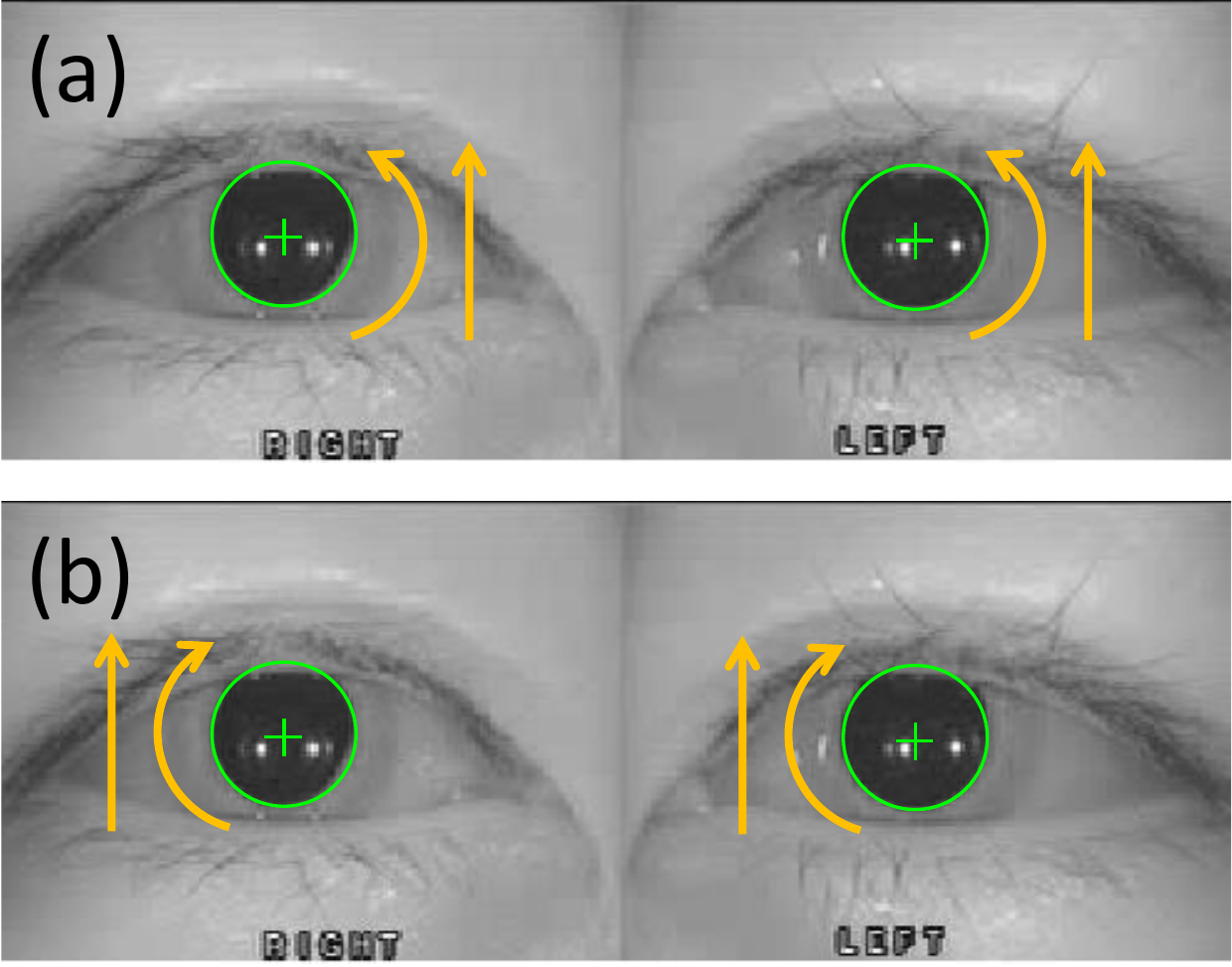}
    	\caption{The posterior canal BPPV types with (a) Rt\_PC\_BPPV specified by a counterclockwise torsion and sometimes accompany with slight up beating, and (b) Lt\_PC\_BPPV type with clockwise torsion and upbeat. These featured movements of the eyes have been learned successfully by the proposed system which is much more robust than the conventional template-matching -based method as in \cite{limec}.}
    	\label{fig:posterior}
    \end{figure}

    \textbf{Posterior canal disorder:} It is specified by the torsional movement of the eye triggered under certain postures of the patient. The posterior BPPVs are categorized into two types: the first one is the ``left posterior canal'' type (Lt\_PC\_BPPV) specified by nystagmus of torsional clockwise along with a slight up-beating when lying with head hanging, and the second one is ``right posterior canal'' type (Rt\_PC\_BPPV) with the nystagmus of torsional counter-clockwise alongside with a slight up-beating when lying with head hanging (shown in Fig. \ref{fig:posterior}). For given video input, we first classify whether it belongs to posterior canal types or not. To classify posterior BPPVs, we use the recurrent neural network with features extracted from a pre-trained CNN for two eyes of the patient at five positions as in Fig. \ref{fig:fig05}: c) d) e) f) g). At each position, the eye's motion is captured by a bidirectional GRU followed by a fully connected layer with 128 neurons. The outputs of five positions are concatenated to a 640-dimension vector and fed into a fully connected layer and a final softmax layer for classification (Fig. \ref{fig:fig04}). 
    
    \textbf{Lateral canal disorder:} BPPV with the horizontal beating of the eyes (shown in Fig. \ref{fig:fig03}) should be distinguished from posterior canal BPPV (PC) which is associated with torsional nystagmus of the eye. Based on the beating direction of the patient eyes corresponding to his/her body posture, it is categorized into geotropic and apogeotropic BPPV (next section). 
    %corrected here
    According to the Barany Society \cite{von2015benign}, the horizontal canal (HC) BPPV is categorized into two different types Canalolithiasis (HC-geo) and Cupulolithiasis (HC-apo) types for each ear (left and right side). So, the total types of HC-BPPV are 4 types: 1) Right ear HC-geotropic (canalolithiasis), 2) Right ear HC-apogeotropic (cupololithiasis), 3) Left ear HC-geotropic (canalolithiasis), and 4) Left ear HC-apogeotropic (cupololithiasis) \cite{von2015benign}. 
    % corrected here
    To classify the different types of geo and apogeo BPPV, LAD extracts the number of beatings and the velocity magnitude of these beatings. And this information is utilized to form a final $20$-dimension features vector for each given video input.

    \subsection{Classifying Geotropic and Apogeotropic Types} % corrected here later--> lateral
    \textbf{Geotropic} BPPV includes: left- and right- lateral canal BPPV, we denote them as ``Lt\_Geo\_BPPV'' and ``Rt\_Geo\_BPPV'', respectively. In the same way, we have the \textbf{apogeotropic} BPPV: ``Lt\_Apogeo\_BPPV'' and ``Rt\_Apogeo\_BPPV''. The definition of these disorders is summarized below. We performed experiments using LSTM or GRU for these types of BPPVs and found that it is not better than random guess indicating that RNN is hard to learn the horizontal beating. All four different types of the lateral canal are characterized by horizontal nystagmus, and end-to-end learning of CNN+RNN did not produce satisfactory classification results. Horizontal nystagmus occurs within a certain moment in time (20s-30s), and ``strong beating'' in each lesion side is the most important indicator in classifying these BPPV disorders (specialists perform diagnosis by looking at particular characteristics of the horizontal nystagmus). In this case, the CNN+RNN model failed to capture the ``strong beating'' signal. The proposed \textbf{Alg. 1} detects strong beating by way of measuring the speed of the beating, which is the most important indicator for classifying different lateral canals.
    
    In the second stream, LAD is learned to extract the different types of features to categorize horizontal beating and classify the beatings into one of four types of horizontal BPPV disorders. 
    
    \subsubsection{Geotropic BPPVs}
    There are two geotropic BPPV types that are defined as follows: Left geotropic canal BPPV composes the geotropic nystagmus and stronger intensity on the left side; while right geotropic canal BPPV includes the geotropic nystagmus and accompanies the stronger intensity on the right side. Both of the cases are considered from the patient's perspective for the postures.
    
    \subsubsection{Apogeotropic BPPVs}
    Two types of Apogeotropic BPPV are differentiated conversely with the geotropic types: Left apogeotropic canal BPPV with apogeotropic nystagmus and stronger intensity on the right side; right apogeotropic canal BPPV with apogeotropic nystagmus and stronger intensity on the left side. 
    
    \subsection{Eye Beating Detection}
    Normally, a video is recorded with the full seven actions of the subject patient with the support of a skillful doctor. In some cases, ``head bending'' may not be included. We detect horizontal beating for acquiring the visual features of the eyes to classify four types of lateral canal BPPV disorder. We can use different available methods to detect and track the eye pupil from the classical techniques that use feature-based methods such as Hough Transform \cite{hog} or the more recent advanced technique PuRe \cite{pure}. Recently, deep learning is an interesting approach for tracking the pupil \cite{deepeye,shahara}. Here we detect eye pupils with a CNN-based network that was inspired by Shaharam \etal \cite{shahara}. 
    % correct here: 
    In our dataset, as described in section \ref{sec:experiments}, to trigger and observe the clear symptom of BPPV disorders, the clinical experimental settings are done in the low light condition to record the videos of the two eyes. In addition, we find that the size and shape of the pupils varied for different patients, and there are cases the eyelid covers the pupil partly or half for which the traditional pupil detector failed to detect correctly the pupils (contains lots of false positives). In practice, we find that the neural network-based CNN detector \cite{shahara} performs more robustly in detecting the pupil compared to the traditional method such as Hough Transform, furthermore, CNN is supported to run on GPU with high speed in processing ($\sim$120 FPS for a single GPU Titan Xp NVIDIA, compared to the traditional detector Hough with $\sim$70 FPS). 
    
    The outputs are the coordinates of the eye $(x,y)$ and the radius of detected pupils $r$. This information of coordinates is utilized to calculate the velocity of the eye movement for beat detection. With each given video, the output should be the beating with the magnitude of beating specified by the velocity magnitude stick to the frame ID that the beating is detected. With each of the positions of the patient, we have the total number of beats left, and right, and the corresponding velocity magnitude. This information is stacked to form the final feature of each video to distinguish four types of BPPVs (Geo and Apogeo types). Algorithm 1 detects the horizontal beating using the coordinates of the eye pupil acquired from every video frame. 
    
    SPV and FPV are computed by Eq. 1. The coordinates of the eye's pupil $ (x,y) $ include $ x $ for the horizontal axis and $ y $ for the vertical axis. For the horizontal canal BPPV, we focus on the horizontal component $ x $ only. Algorithm 1 illustrates how to detect one beating that occurred with the coordinate $ x $ of the eye. The beating velocity is computed by the following equation (Fig. \ref{fig:fig06}):
    \begin{equation}
    v = \frac{\Delta x}{\Delta t},
    \end{equation}
    where $ \Delta x $ is the distance the eye travels within the space of the video frame during the time step $ \Delta t $. 
    The beatings for two eyes of the patient are scattered in Fig. \ref{fig:fig01_right} and Fig. \ref{fig:fig01}, the behavior of both eyes is almost the same. In some cases, if one of the eyes closes then we can still get the information from the other one. Fig \ref{fig:fig01_right} shows the horizontal beating over time for a video of a patient who is positive with left lateral canal BPPV. 
    
    In the region from frame 1900 to frame 2200 when the patient turned his head right, the right beating is dominated (red points). A few moments later, when he turned left (frame 2500 to frame 3000), the left beating appeared (blue points). 
    
    \begin{table*}
    \caption{Classifying the torsional nystagmus (posterior canal types)}
    \label{tab:table23}
    \begin{center}
        \begin{tabular}{c c c c}
        \hline
        Reference & Feature extraction & Classifier & Be able to learn End-to-End \\
        \hline\hline
        Lim \etal \cite{limec} & Hand-crafted features (Template Matching) & CNN & \xmark \\
        The proposed LAD & CNN (pre-trained on ImageNet dataset) & CNN+BiGRU & \checkmark \\
        \hline
        \end{tabular}
    \end{center}
    \end{table*}

    \begin{table*}
    \caption{Classifying the horizontal nystagmus (lateral canal types).}
    \label{tab:table24}
    \begin{center}
        \begin{tabular}{ c c c c c }
        \hline
        Reference & Feature extraction & Eye tracking method & Classifier & Detect beating \\
        \hline\hline
        Lim \etal \cite{limec} & Sum of the transient velocity (all frames) & Hough Transform & CNN & \xmark \\
        The proposed LAD & Velocity and count of detected beatings (filtered frames with no beating) & CNN & SVM & \checkmark \\
        \hline
        \end{tabular}
    \end{center}
    \end{table*}
    
    \begin{table*}
    \caption{Comparison of the recent works in analysis of nystagmus. The performance is reported based on the accompanying dataset. The abbreviations are as follows: PCA: Principle Component Analysis, MNN: Multilayer Neural Network, VNG: Video-nystagmography, FLD: Fisher Linear Discriminant, CNN: Convolutional Neural Network, VOG: Video-oculography, F-CNN: Fully Convolutional Neural Network, BiGRU: Bi-directional Gate Recurrent Unit. } % corrected here
    \label{tab:table22}
    \begin{center}
        \begin{tabular}{ c c c c c c c }
        \hline
        Reference & Published year & Feature dataset & Classifier & Classes & Body pose video & Overall accuracy. \\
        \hline\hline
        Ben \etal \cite{slama} & 2017 & Temporal, frequency & PCA-MNN & 3 & \xmark & - \\ %95.5\%
        Ben \etal \cite{slama3} & 2018 & Clinical VNG features & FLD-MNN & 2 & \xmark & - \\ %91.58\%
        Lim \etal \cite{limec} & 2019 & Eye tracking video & Template matching + CNN & 8 & \xmark & 79.8 \% \\ %
        Ben \etal \cite{slama4} & 2019 & Nystagmus signal & CNN & 3 & \xmark & - \\ %96.36\%
        Yiu \etal \cite{yiuhoi} & 2019 & VOG sequence & F-CNN & 2 & \xmark & - \\ %95.8\%
        Ben \etal \cite{slama2} & 2020 & VNG sequence & FLD-MNN & 2 & \xmark & - \\ %94.73\%
        \hline
        \textbf{The proposed LAD} & 2022 & Eye tracking, body video & Hyrid CNN+BiGRU & 6 & \checkmark & \textbf{91.0}\%\\
        \hline
        \end{tabular}
    \end{center}
    \end{table*}
    \subsection{Features from detected beating} 
    
    For each left and right pair of beating, a softmax (Eq. 2) is applied to normalize the raw beating number to a range between $0$ and $1$. The pair of average beating velocity magnitudes are also normalized in the same way. For horizontal canal disorder, we care for the beating on the positions in Fig. \ref{fig:fig05} $d, e, f, g$. \textit{With four positions of the patient, we have a total of \textbf{eight} normalized values of the left and right beating numbers}. The binary value $ 0 $ for the left side and $ 1 $ for the right side, denotes the side that has the maximum total beating, and performed a softmax normalization for the pair of the max total beating on the left side and the max total beating on the right side as the following equation:
    \begin{equation}
    \text{Softmax}(\mathbf{x})_i = \frac{\exp(x_i)}{\sum_{j=1}^{2}\exp(x_j)},
    \end{equation}
    where $ i\in {1,2} $ and $ \mathbf{x}=(x_1,x_2)\in \mathbb{R}^2 $. After this step, we have three more features for positions $d,e$. Now apply the same way for the average beating velocity for both sides, we have three more features for positions $d,e$ (lateral head turn test). Similarly, \textit{six features are extracted for the position} $f,g$ when the patient lying on the bed (with the head hanging left and right with the Dix-Hallpike maneuver). For the position of head bending (Fig. \ref{fig:fig05}\textit{b}), and lying down (Fig. \ref{fig:fig05}\textit{c}), the beating direction is important, the magnitude of beating is not considered when diagnosing lateral canal BPPV types. 
    
    Finally, \textit{we form a $20$-dimension vector extracted from a video that contains information on the horizontal beating of both affected sides of the patient}. These feature vectors are used as input to a classifier such as a deep neural network or an SVM for classification. The experimental results and analysis are described in the next section. We also conducted the experiments with the method in \cite{limec} that is most relevant to our work on the nystagmus analysis for BPPV.

\section{Experiments}
    \label{sec:experiments}
    \subsection{Data Collection}
    The videos were recorded in \textit{*.avi} format with the resolution of $ 240 \times 320 $ and at 30 FPS, each video includes three sub-videos: left eye, right eye, and whole patient's body scene. As shown in \cite{bronstein}, to observe the symptom of the eyes clearly, the diagnostic tests were conducted in a dark environment. There are two popular positional tests that are used generally to diagnose BPPV disorder which are described in more detail below.
    % corrected here
    The room is set up with a low light condition to trigger BPPV symptoms in patients. The patient is staying on the bed. There is one doctor standing by side to perform Dix-hallpike maneuvers and lateral canal tests. There is one camera located 2.3m far from the patient to capture the whole scene of both patient and doctor.
    
    The first test is the Dix-Hallpike maneuver which is used to diagnose the type of posterior canal BPPV \cite{tangheng}. This test is performed independently for both the left and right sides. For the left side, the patient (equipped with a goggles camera attached to his head) is seated straight on the bed and a doctor will help him to turn his head $ 45^\text{o} $ to the left and quickly lie down so that the patient's head is hanging left out of the bed (Fig. \ref{fig:fig05}\textit{f}). The right side test is similar to the left side but now the head of the patient is turned $ 45^\text{o} $ to the right before lying down in a hanging position (Fig. \ref{fig:fig05}\textit{g}). 
    
    Keeping each side test duration for less than thirty seconds and the eyes of the patient are observed for any nystagmus that occurred during this period. In the Dix-Hallpike test, the vertical-torsional nystagmus is often observed either clockwise or counter-clockwise direction which specified the left posterior canal or right posterior canal BPPV type, respectively.
    \begin{figure}[!ht]
    	\centering
    	\includegraphics[width=1.0\linewidth]{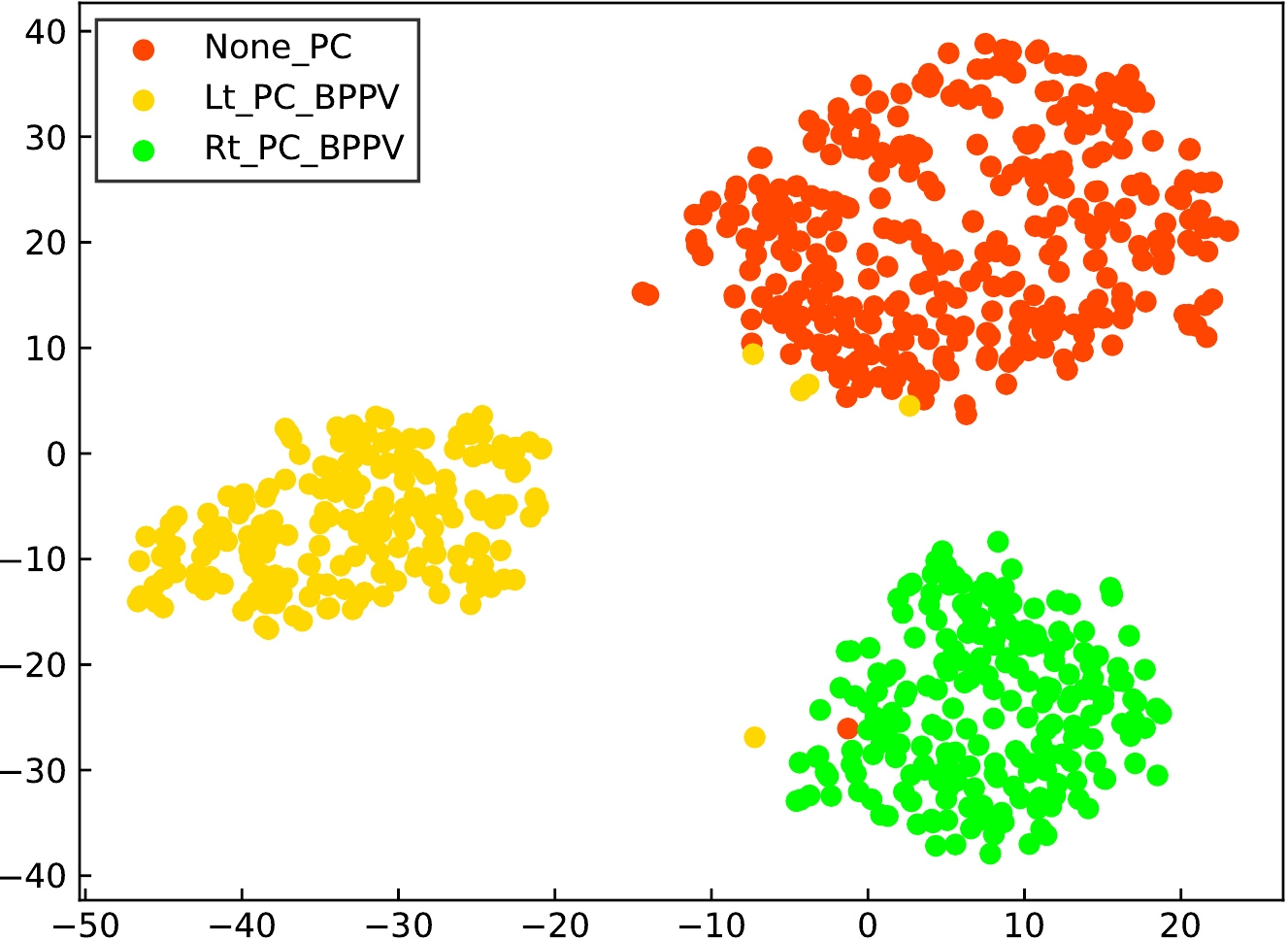}
    	\caption{t-SNE plot for the first round, classifying video into three kinds of BPPV groups: Lt\_PC\_BPPV, Rt\_PC\_BPPV, and None\_PC. The plot is scattered into the 2D plane for 746 data points (the whole dataset). It is best viewed in color.}
    	\label{fig:fig09}
    \end{figure}
    
    The second test performed on the data collection is the lateral canal test \cite{bhatta}. This test is done by keeping the patient lying (head is on the bed) with the eyes looking straight to the ceiling and conducting two maneuvers. For the left side, quickly turn the patient's head toward the left (Fig. \ref{fig:fig05}\textit{d}) For the right side, turn the patient's head quickly to the right (Fig. \ref{fig:fig05}\textit{e}). After each of two of these maneuvers, the eyes of the patient are observed within less than thirty seconds to see if any horizontal nystagmus has occurred during this period. 
    
    In the lateral canal test, four types of BPPV are clearly observed including left and right geotropic lateral canal BPPV and left and right apogeotropic lateral canal BPPV. The patient's body (captured by a remote camera) and two eyes (captured by goggles) were performed simultaneously and stacked to a video with three sub-videos for the left eye, right eye, and the body (to see the scene later). 
    Each video belongs to a unique patient, ranging from 10 to 70 yrs old, including both genders. 
    
    The two physical tests are consecutively recorded resulting in a unified video with a duration of approximately three minutes and a size of 15-20 MB on average. % corrected here
    By observing the behaviors of the patient eyes in all positions we tested above, two otology experts in a hospital analyze and classify which kind of disorder the patient belongs to.     
    The recorded videos with subtle symptoms, low quality, or incomplete have been removed from the cohort. A total of 746 data from patients with six classes of BPPV disorder with 406 data for posterior canal types and 340 data for lateral canal types were used in the experiments. 
    
    Table \ref{tab:table4} shows that posterior canal BPPV is more common and accounts for a proportion of 26\% and 29\% for the left and right types, respectively. Table \ref{tab:table22} illustrates a comparison of different works on various data collections in BPPV analysis. Our dataset is different from previous datasets: it includes movements of both eyes that are aligned with the different postures of the patient (following the Dix-Hallpike test procedure).
    
    \subsection{Training Details}
    A pre-trained CNN ResNeXt32x48d \cite{fixres} is used as a feature extractor for both spatial and temporal streams which results in a 2048-dimension vector for each frame of the two eyes. We tried with another feature extractor Inception V3, however, the recognition accuracy is behind 5\% compared to the ResNeXt32x48d. In the part of classification for posterior BPPVs, we utilized a bidirectional GRU to learn the eye motions in five corresponding postures of the patient. 
    
    In this part, each GRU network outputs a vector of 128 dimensions and is stacked to form a final 640-dimensions vector. A fully connected layer (512 neurons) and a softmax layer (3 neurons) were added to classify the given video into one of 3 classes: None\_PC, ``Lt\_PC\_BPPV'' and ``Rt\_PC\_BPPV''. All of these networks used ADAM optimizer \cite{kingma} with default hyper-parameters for training and cross-entropy loss for classification. The learning rate was set to 0.001. 
    
    In the classification part for four types of BPPVs, each video was extracted to a $20$-dimension vector and then fed into a linear-kernel SVM classifier for classification \cite{svm}. The use of SVM was a designer's choice as SVM performed similarly to FC+softmax and SVM was a simpler choice requiring less memory.
    
    \begin{table}[htbp]
    \caption{The collected data was split into train and test sets with a ratio of 90\% and 10\%, respectively. ``Per class rate'' refers to the proportion of data for each class in the whole data.}
    \label{tab:table4}
    \begin{center}
        \begin{tabular}{ c c c c }
        \hline
        Disorder types & \# Train sample & \# Valid sample & Per class rate\\
        \hline\hline
        Lt\_Geo\_BPPV & 73 & 9 & 11 \% \\
        Rt\_Geo\_BPPV & 78 & 9 & 12 \% \\
        Lt\_Apo\_BPPV & 84 & 10 & 13 \% \\
        Rt\_Apo\_BPPV & 69 & 8 & 10 \% \\
        Lt\_PC\_BPPV & 173 & 20 & 26 \% \\
        Rt\_PC\_BPPV & 191 & 22 & 29 \% \\
        \hline
        Total & 668 & 78 & 100 \% \\
        \hline
        \end{tabular}
    \end{center}
    \end{table}
    
    \subsection{Evaluation metrics}
    To evaluate performance, the following metrics were used: (1) accuracy, (2) precision, (3) recall, and (4) F1-score. Accuracy is computed by the proportion of correct predicted class over total data. Precision shows the proportion of predicted data that are correctly classified. Recall gives the proportion of actual samples that are correctly classified. 
    
    The F1-score determines the performance of the model to classify all data, which makes a balance between precision and recall and gets its best value at 1 and worst score at 0. These standard measures are computed and presented by the following standard equations:
    \begin{gather}
    %\begin{align}
        \text{Accuracy} = \frac{\text{TP + TN}}{\text{TP + FP + TN + FN}}\\
        \text{Precision} = \frac{\text{TP}}{\text{TP + FP}}\\
        \text{Recall} = \frac{\text{TP}}{\text{TP + FN}}\\
        \text{F1 score} = 2\times \frac{\text{Precision} \times \text{Recall}}{\text{Precision + Recall}},
    %\end{align}
    \end{gather}
    where TP, TN, FP, and FN denote the True Positive, True Negative, False Positive, and False Negative, respectively. In the total 746 data samples, train and valid set are split randomly with a ratio of 90\%/10\% three times. 
    
    The aforementioned metrics are calculated for three validation sets and averaged. The split procedure follows that used in obtaining the splits for JHMDB/ HMDB dataset for action recognition \cite{jhuangiccv2013,Kuhneiccv2011}. Instead of a 7:3 split of HMDB, we used a 9:1 split for the evaluation.
    \begin{table}[t]
    \caption{Input data have been roughly categorized into three groups using BiGRU: Not posterior canal (None\_PC\_BPPV), left posterior canal (Lt\_PC\_BPPV), and right posterior canal BPPV (Rt\_PC\_BPPV). It is performed on three different splits validation sets and takes the average. Then, the None\_PC\_BPPV is classified into the rest of the four classes with lateral canal BPPV types.}
    \label{tab:table1}
    \begin{center}
        \begin{tabular}{ c c c }
        \hline
        BPPV types after the $1^{st}$ stage & Accuracy & Number of samples \\
        \hline\hline
        Lt\_PC\_BPPV & 0.90 & 20 \\
        Rt\_PC\_BPPV & 0.94 & 22 \\
        None\_PC\_BPPV & 0.94 & 36 \\
        \hline
        \end{tabular}
    \end{center}
    \end{table}
    
    \subsection{Results}
\begin{figure}[!ht]
    	\centering
    	\includegraphics[width=1.0\linewidth]{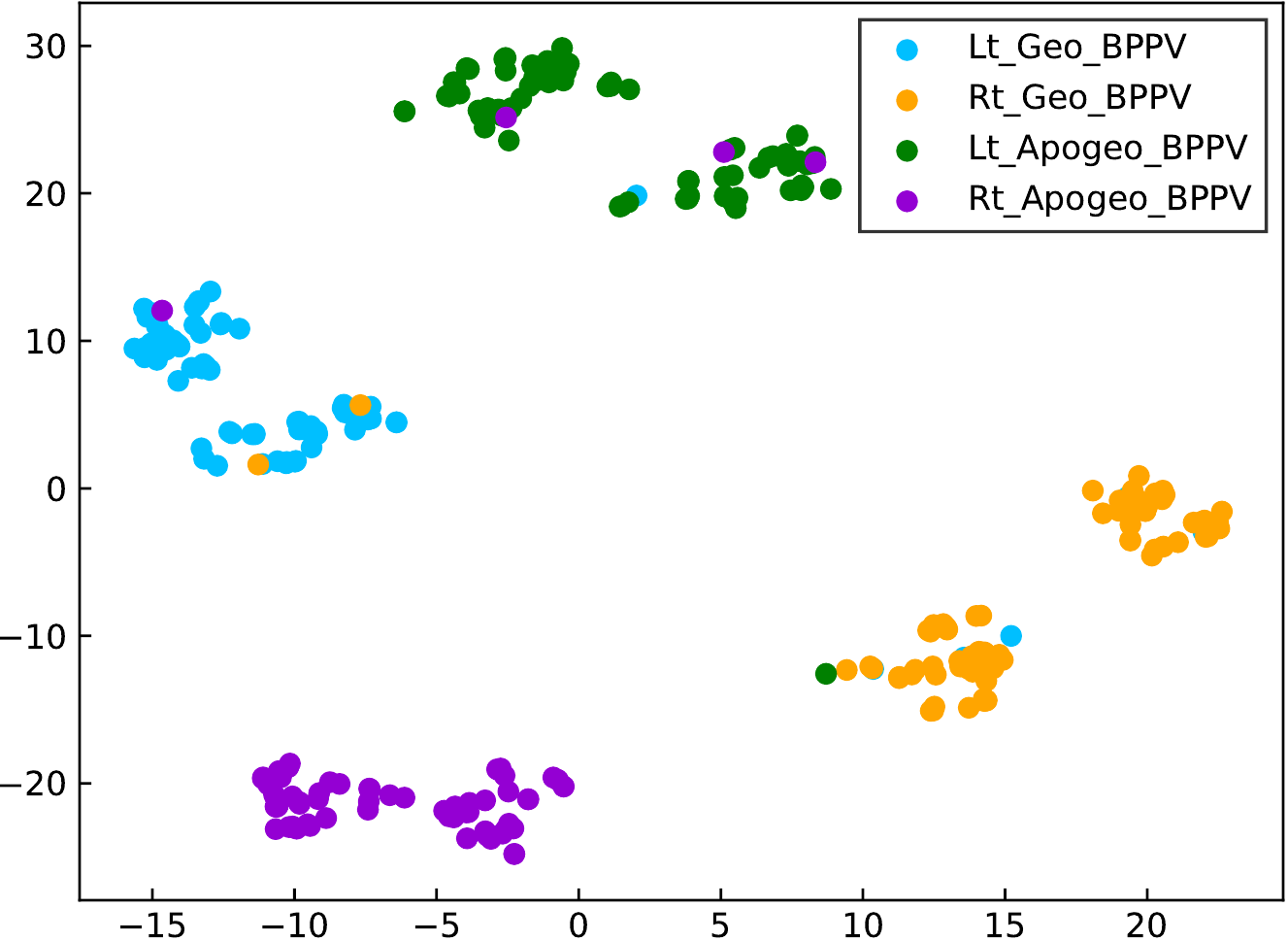}
    	\caption{t-SNE plot for 4 kinds of BPPV disorder: geotropic (left, right) and apogeotropic (left, right) classes (plotting for a total of 340 data points for lateral canal types). It is best viewed in color.}
    	\label{fig:fig02_2}
    \end{figure}

    We show that the four lateral canal types of Geotropic and Apogeotropic BPPV disorder can be well classified based on the proposed method. Fig. \ref{fig:fig02_2} visualizes the feature vectors of lateral BPPV types in 2D plane with t-SNE visualization \cite{laurens}. It clearly shows that these vectors are clearly separable and belong to four separate groups. 
    
    These feature vectors are constructed by the measurement of horizontal beating velocity alongside magnitude and the beating intensity which are the crucial clues for distinguishing the horizontal canal BPPVs. Table \ref{tab:table3} shows the validation accuracy of 95\% for lateral canal type, 93\% for posterior canal types, and 91\% for overall. For the posterior canal BPPV, the classification results are shown in Fig. \ref{fig:fig09}, Table \ref{tab:table3}, and Table \ref{tab:table1} with three classes: Lt\_PC\_BPPV, Rt\_PC\_BPPV, and None\_PC\_BPPV, showing that the BPPV types with rotational features are well extracted. 
    \begin{table}[!ht]
    \caption{Accuracy of two stages and the final process. Step \textit{1}, data are categorized into posterior canal types (Lt\_PC\_BPPV, Rt\_PV\_BPPV) and None\_PC\_BPPV. Step \textit{2}, the None\_PC\_BPPV data are classified into four types of lateral canal BPPVs. And finally, the accuracy for the final predicted labels is calculated.}
    \label{tab:table3}
    \begin{center}
        \begin{tabular}{ c c c c }
        \hline
        Type & Posterior type & Lateral type & Overall accuracy\\
        \hline\hline
        Accuracy & 0.93 & 0.95 & 0.91 \\
        \hline
        Number of videos & 42 & 36 & 78 \\
        \hline
        \end{tabular}
    \end{center}
    \end{table}
    
    To visualize how the ``gated recurrent unit'' (GRU) differentiates different posterior canal BPPV types, the features from the top layer of the model in Fig. \ref{fig:fig04} are extracted. The classification results and t-SNE plot demonstrate that the aforementioned featured motion of the eyes has been successfully captured by the bidirectional GRU model (BiGRU) to distinguish posterior canal types (torsional movements) and lateral canal types (horizontal beatings). Overall, the whole system classifies the videos in the validation set with 0.91, 0.90, and 0.90 for precision, recall, and F1 scores as shown in Table \ref{tab:table2}, respectively.
    
    Table \ref{tab:table5} compares the baselines and our method. It shows that method in \cite{limec} has an accuracy of 48.7\% which is much lower than that of our \textbf{91.0\%} on our dataset. Low resolution, low-lighting conditions, and partially obscured iris make it a challenging problem, and the simple template-matching based-method is not robust enough to capture the torsion in the iris. It also shows that our algorithm is far superior in comparison to the temple-matching method \cite{limec} which can never be robust to sporadic eyelid closure. 
    
    \textbf{Alg. 1} detects frames with beating, which is important information to specialists for diagnosing lateral canal BPPV, it works as the attention mechanism filters out all frames with no beating, and the method in \cite{limec} doesn't have this capability. The pupil detection is performed independently to mimic how experts are performing the beating (nystagmus) detection.
    \begin{table}[!ht]
    \caption{Comparison of the baselines and the proposed method in terms of validation accuracy in our dataset, $^\dagger$ indicates that we implemented the method \cite{limec} in our dataset. Non-Information Rate (NIR) indicates the most proportional class rate in the collected dataset.}
    \vspace{-15pt}
    \label{tab:table5}
    \begin{center}
        \begin{tabular}{ c c }
        \hline
        Method & Overall Accuracy (\%) \\
        \hline\hline
        NIR (baseline) & 29.0 \\
        Lim \etal \cite{limec}$^\dagger$ & 48.7 \\
        \textbf{Proposed LAD} & \textbf{91.0} \\
        \hline
        \end{tabular}
    \end{center}
    \end{table}
    \vspace{-15pt}
    Table \ref{tab:table22} shows the recent works in BPPV research. Each reference has a separate dataset and a different number of classes for nystagmus analysis. The most related data to our work is the work of Lim \etal \cite{limec}. They collected data from ten different positions of the patient, however, they failed to apply RNN to classify the disorders and used the handcrafted feature as the feature descriptor. In contrast, our dataset contains 7 postures of the patient, applied successfully an RNN (with a bidirectional GRU) for classification of BPPV disorders, and proposed an algorithm that detects the horizontal beating that can filter out all irrelevant frames that have no beating, which results in the more accurate information for classifying the torsional movements as well as the horizontal nystagmus in the given video input. 
    \begin{table}[!ht]
    \caption{The classification results for six classes of BPPV disorders. The results are performed on three different splits validation sets and are taken average. LC stands for ``lateral canal'' while PC stands for ``posterior canal''.}
    \vspace{-15pt}
    \label{tab:table2}
    \begin{center}
        \resizebox{1.0\hsize}{!}{
        \begin{tabular}{ c c c c c c }
        \hline
        Type & Disorder type & Precision & Recall & F1 score & \# Test samples \\
        \hline\hline
        \multirow{4}{*}{LC}
        &Lt\_Geo\_BPPV & 0.92 & 0.85 & 0.88 & 9 \\
        &Rt\_Geo\_BPPV & 0.81 & 0.89 & 0.84 & 9 \\
        &Lt\_Apo\_BPPV & 0.94 & 0.97 & 0.95 & 10 \\
        &Rt\_Apo\_BPPV & 0.96 & 0.88 & 0.91 & 8 \\
        \hline
        \multirow{2}{*}{PC}
        &Lt\_PC\_BPPV & 0.92 & 0.90 & 0.91 & 20 \\
        &Rt\_PC\_BPPV & 0.93 & 0.94 & 0.93 & 22 \\
        \hline
        % \diagbox[width=7.2em]{ }{ } & Average / Total & 0.91 & 0.90 & 0.90 & 78 \\
         & Average / Total & 0.91 & 0.90 & 0.90 & 78 \\
        \hline
        \end{tabular} }
    \end{center}
    \end{table}
    \vspace{-15pt}
\subsection{Learned Feature Visualization}
\begin{figure}[!ht]
    	\centering
    	\includegraphics[width=1\linewidth]{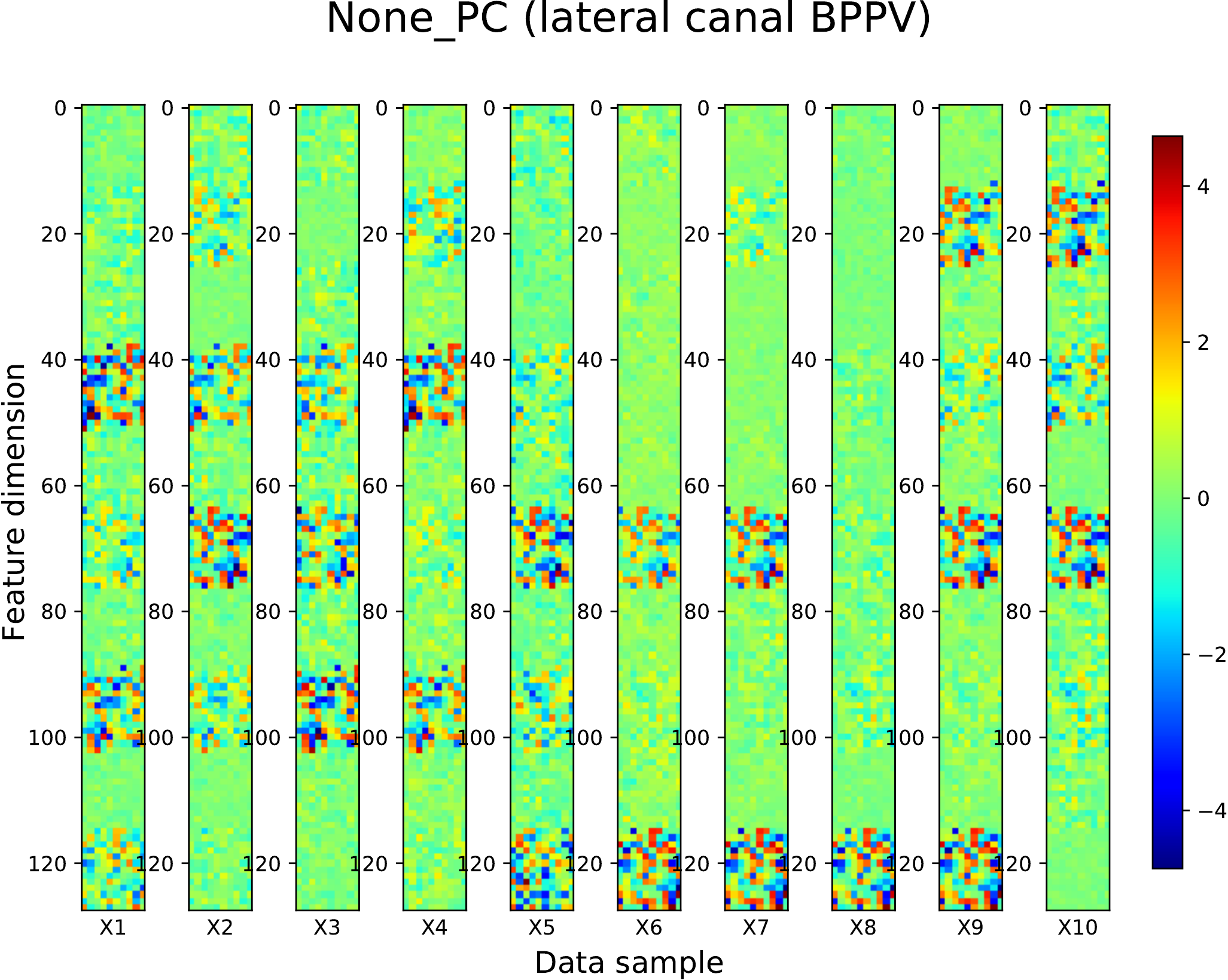}
    	\caption{Visualizing the learned features of the None\_PC class.}
    	\label{fig:nonepc}
    \end{figure}

\begin{figure}[!ht]
    	\centering
    	\includegraphics[width=1\linewidth]{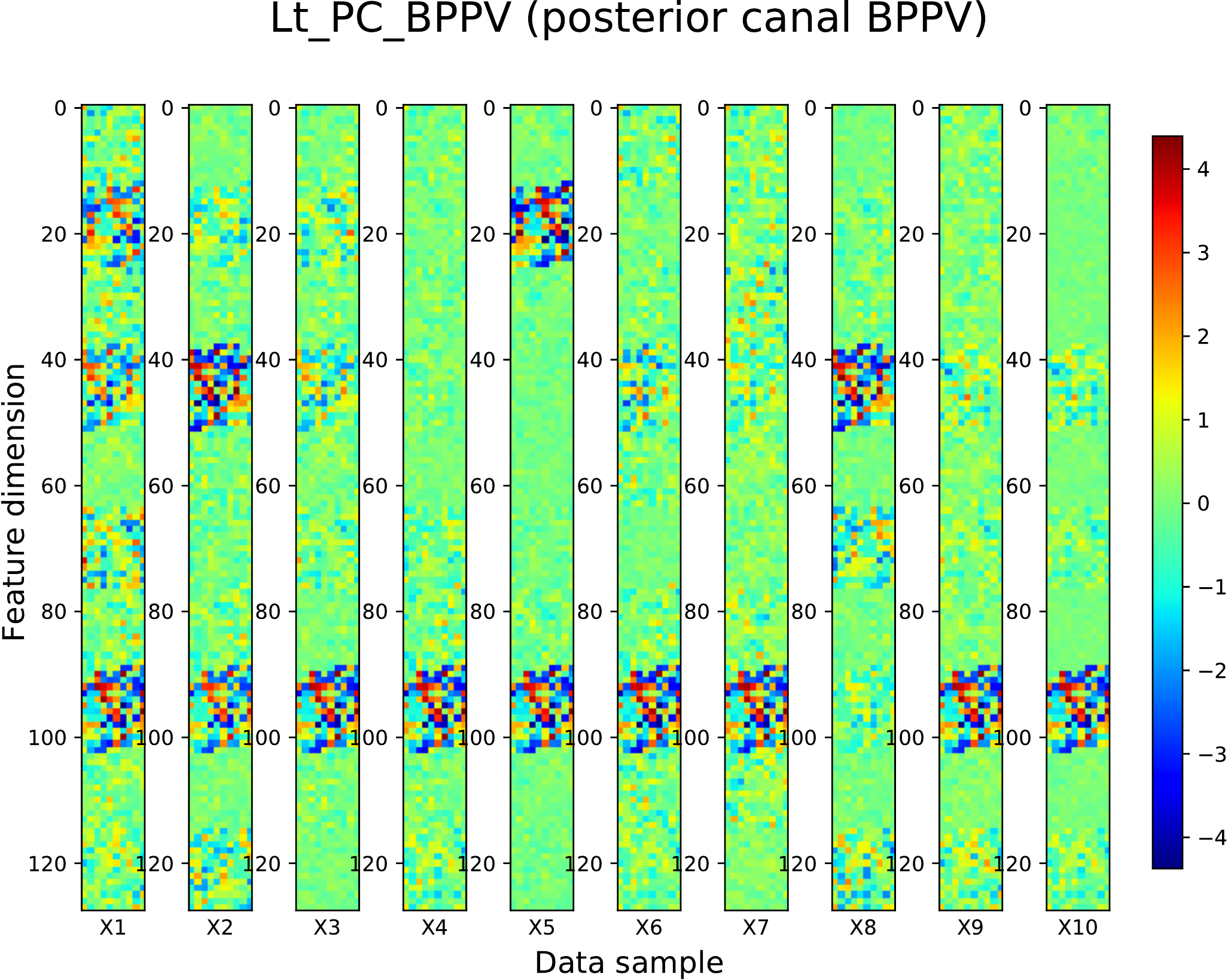}
    	\caption{Visualizing the learned features of the Lt\_PC\_BPPV class.}
    	\label{fig:ltpc}
    \end{figure}

\begin{figure}[!ht]
    	\centering
    	\includegraphics[width=1\linewidth]{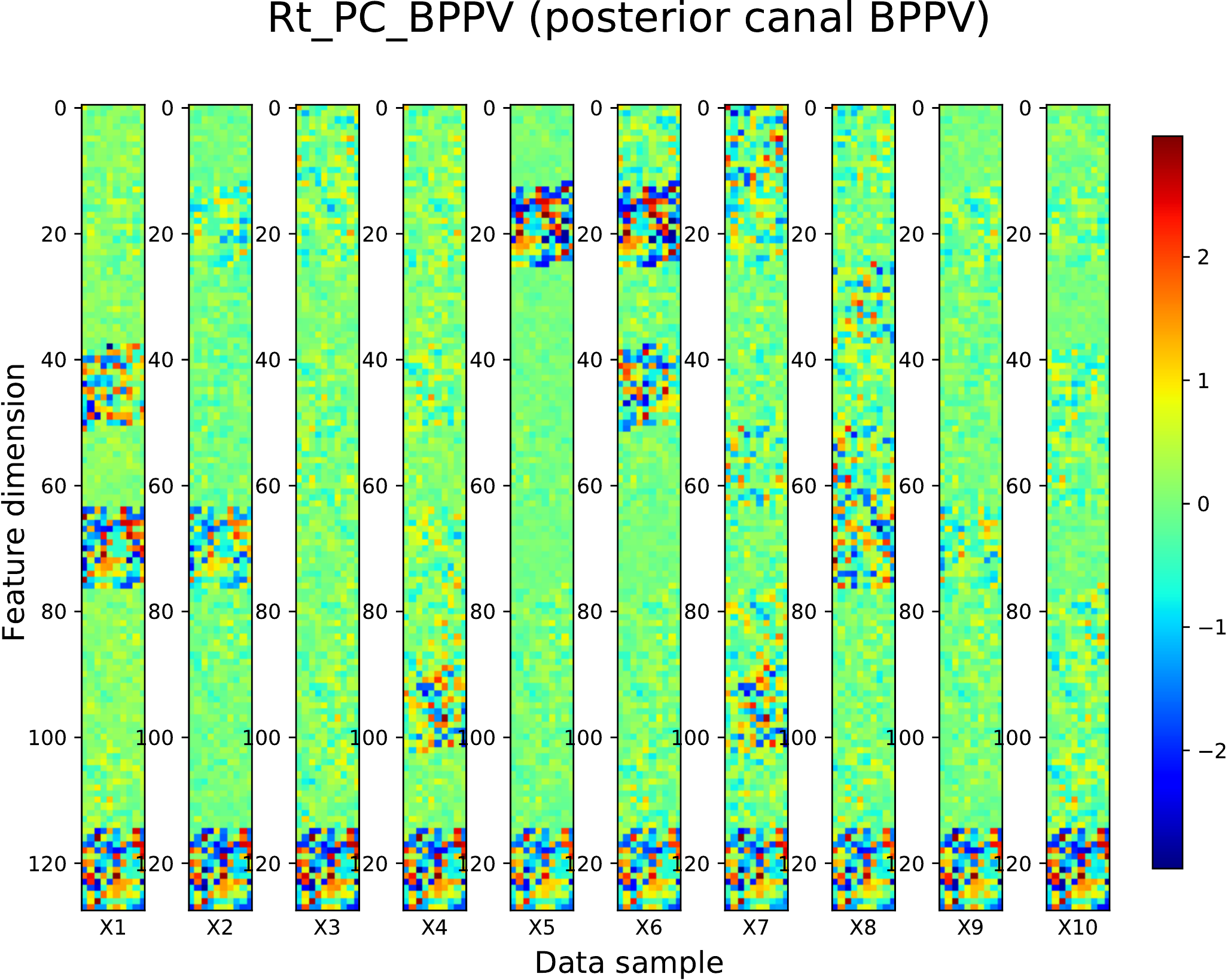}
    	\caption{Visualizing the learned features of the Rt\_PC\_BPPV class.}
    	\label{fig:rtpc}
    \end{figure}

    Fig. \ref{fig:nonepc}, \ref{fig:ltpc}, and \ref{fig:rtpc} visualize the features of ten different samples from three classes that have been learned by using the top layer that is trained in the proposed model. The top layer of the proposed model produces 1280-dimensions feature vectors that have been reshaped to 2D feature maps ($10\times 128$) for visualization. 
    It shows that class ``None\_PC'', ``Lt\_PC\_BPPV'' and ``Rt\_PC\_BPPV'' have been extracted with the discriminative representations, indicating that the proposed model with RNN network has successfully learned the behaviors of torsional (Lt\_PC\_BPPV and Rt\_PC\_BPPV) and horizontal (None\_PC) movements of the eyes. These visualizations also strengthen the t-SNE plot in Fig \ref{fig:fig09} that three BPPV types have been well clustered. 
    
\begin{figure}[!ht]
    	\centering
    	\includegraphics[width=1.0\linewidth]{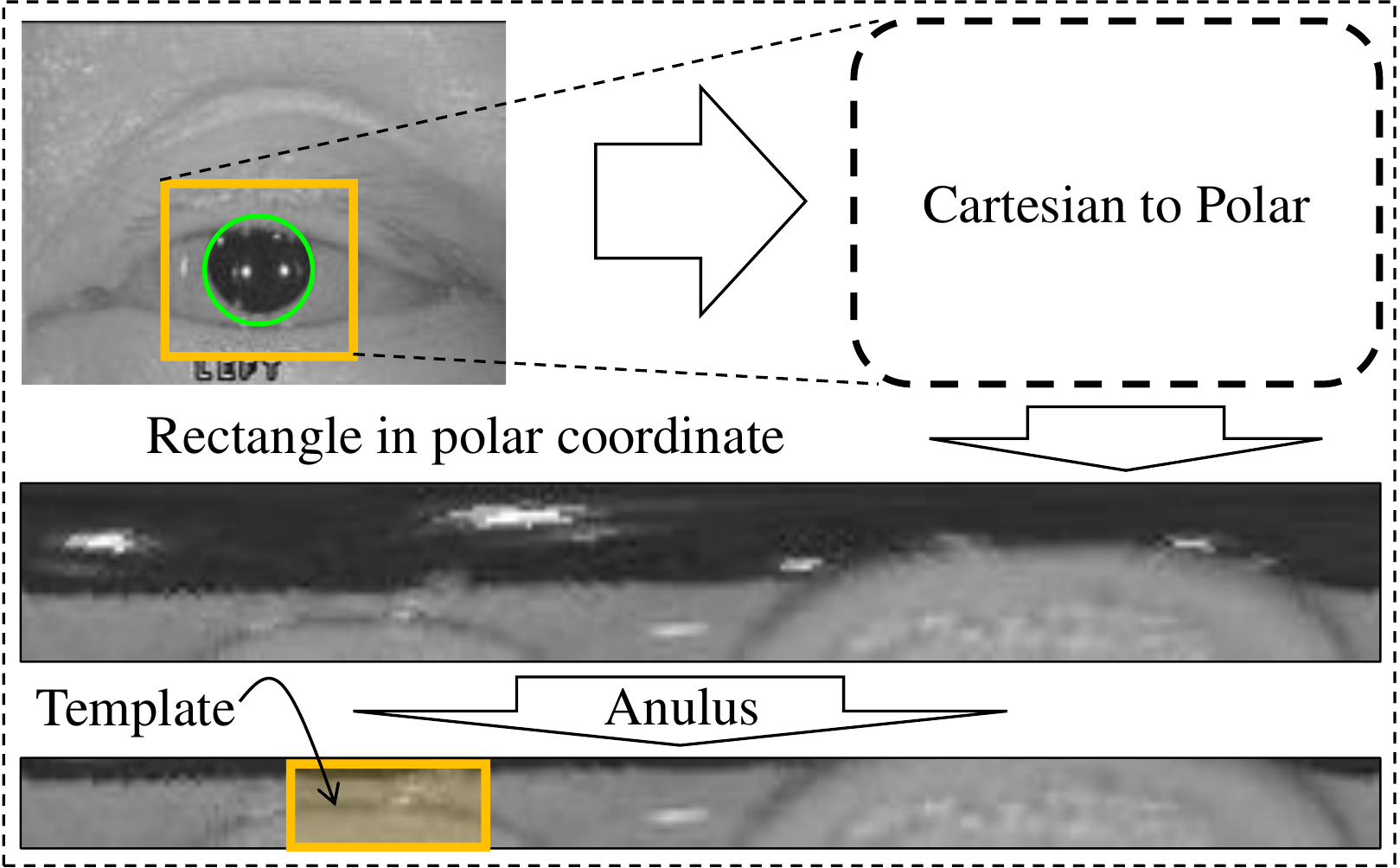}
    	\caption{The process to covert an annulus in Cartesian coordinate to a rectangle in a polar coordinate that is described as in method \cite{limec}. }
    	\label{fig:limec}
    \end{figure}

    It is important to categorize well the video input in the first stage so that the next stage can use its results for the next stage to diagnose the rest of the BPPV types. For comparison with the full features-based method in \cite{limec} that is closest to our work, we reimplemented that technique with template-matching for torsional categorization in our dataset (Fig \ref{fig:limec}). The comparison results are reported in Table \ref{tab:table5}.
    
\section{Conclusion}
    \label{sec:conclusion}
    This paper proposes a deep learning-based system to support the doctor in automatically diagnosing BPPV disorder effectively based on the visual motions of the patient eyes in the standard maneuver tests. Abnormal eye behaviors are triggered and these features are captured and classified into six different types of BPPV. The torsional movements of the eyes are successfully captured by a BiGRU network, while the horizontal nystagmus is distinguished by the normalized beating feature that is extracted from the eye pupil's horizontal coordinate. 
    
    By detecting accurately the horizontal beating that occurred in a specific timestamp of the whole video, the final 
    features of the input video are extracted to disorder types. A two-stream deep architecture network is constructed to extract different types of features for accurate classification. The proposed method can successfully classify disorders with an accuracy of 93\% for the \textit{Posterior Canal} and 95\% for the \textit{Geo} and \textit{Apogeo} BPPV, respectively.

\section{Discussion and Future Works}
    \label{discussion}
    Our work focuses on the application of artificial intelligence to investigate the BPPV disorders in medical data which can be observed by visual movements in the recorded videos. This can aid the doctors significantly in the context that the number of patients increases and hence the video data collected arose every year. The recent advances in deep learning such as CNN demonstrate the very powerful capabilities in learning the image and video representation in the carefully collected dataset \cite{krizhevsky2012imagenet} which is also proved to be beneficial in the custom dataset as in our work. 

    % corrected here, our --> Our
    Our work has some limitations such as the feature of the patient body when integrates into the eyes movement features do not give the desired performance. Therefore, currently, we use labels that mark the exact posture of the patient for each moment in the video. To make the whole process a fully automatic diagnosis, the feature extractor on the human body should extract more accurate subtle details like when the head turns left or right which is crucial in diagnosing BPPV disorders.
    
    The more recent framework Vision Transformer (ViT) \cite{dosovitskiy2021an} has emerged as a promising approach that outperforms CNN in some tasks in visual recognition. It is a potential backbone for improving the BPPV disorders automatic process. ViT is a powerful model but heavy computation, especially in the video dataset. We let it for future work.

\section*{Acknowledgment}
    The authors would like to thank Jaewon Lee, Minseong Yoon, and Jeeeun You for their helpful advice and strong support during the data collection and reviewing of the manuscript. Also thanks for the great support from doctor Choi from Chungnam National Hospital University for providing the data and performing data checking as well as doing labeling for the all different types of BPPV disorders.

\vspace{-20pt}
\begin{IEEEbiography}[{\includegraphics[width=1in,height=1.25in,clip,keepaspectratio]{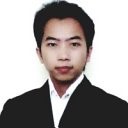}}]{Pham Xuan Trung} received his B.S. degree in the School of Electronics and Telecommunications (SET) at Hanoi University of Science and Technology (HUST) in 2014. He is currently working toward his Ph.D. at KAIST under the supervision of Prof. Chang D. Yoo. His doctoral research interests include Speech Processing, Self-Supervised Learning, and Computer Vision.
\end{IEEEbiography}

\vspace{-20pt}
\begin{IEEEbiography}[{\includegraphics[width=1in,height=1.25in,clip,keepaspectratio]{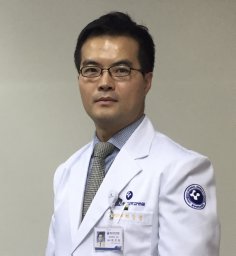}}]{Jin Woong Choi}
Associate professor Choi (Otorhinolaryngology-Head and Neck Surgery) received his Ph.D. at the Department of Otorhinolaryngology-Head and Neck Surgery of Chungnam National University, School of Medicine 2015. His interest includes Otology and neurotology.
\end{IEEEbiography}

\begin{IEEEbiography}[{\includegraphics[width=1in,height=1.25in,clip,keepaspectratio]{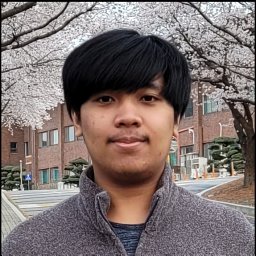}}]{Rusty John Lloyd Mina} was an M.S. student and a Machine Learning Researcher at KAIST from 2019 to 2021. Prior to that, he received his B.S. degree in Electronics and Communications Engineering at the University of the Philippines - Diliman in 2018. His research interests include ML Art, Computer Vision, and Continual Learning.
\end{IEEEbiography}

\begin{IEEEbiography}[{\includegraphics[width=0.8in,height=0.8in,clip,keepaspectratio]{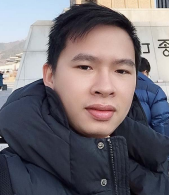}}]{Thanh Nguyen} received the B.S. degree in electronic and automation engineering from the Ho Chi Minh City University of Science and Technology, in 2015. He is currently pursuing his M.Sc. and Ph.D. degrees with the Korea Advanced Institute of Science and Technology. His research interests include machine learning, deep learning, and reinforcement learning.
\end{IEEEbiography}

\begin{IEEEbiography}[{\includegraphics[width=1in,height=1.25in,clip,keepaspectratio]{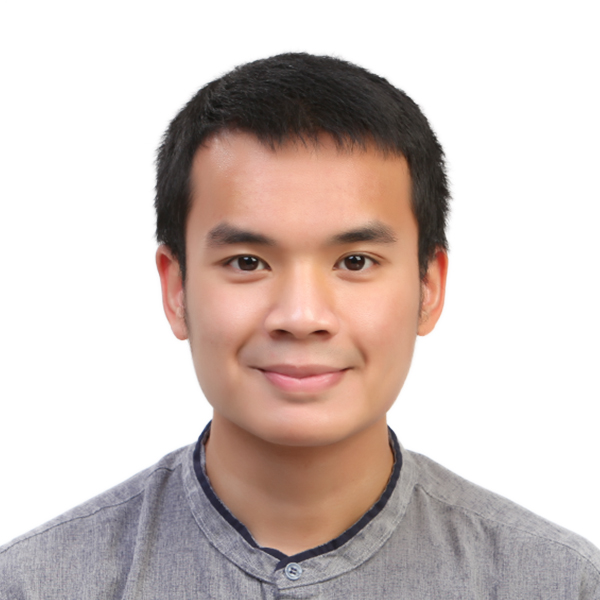}}]{Sultan Rizky Madjid} was an M.S. student and a Machine Learning Researcher at KAIST from 2021 to 2023. Prior to that, he received his B.S. degree in Electrical Engineering with a double major at Mechanical Engineering at KAIST in 2021. His research interests include Model Compression, Sparse Representations in Deep Learning, and Continual Learning.
\end{IEEEbiography}

\begin{IEEEbiography}[{\includegraphics[width=1in,height=1.25in,clip,keepaspectratio]{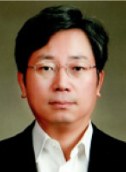}}]{Prof. Chang D. Yoo} received the B.S. degree in Engineering and Applied Science from the California Institute of Technology, the M.S. degree in Electrical Engineering from Cornell University, and the Ph.D. degree in Electrical Engineering from MIT. His current research interest is machine learning, signal processing, computer vision, and audio processing. He is a Member of Tau Beta Pi and Sigma Xi. He is the director of the Video Turing Test Research Center and also the AI Fairness Research Center.
\end{IEEEbiography}

\bibliographystyle{IEEEtran}
\bibliography{egbib}

\EOD

\end{document}